\def\year{2020}\relax
\documentclass{article} 
\usepackage{aaai20}  
\usepackage{times}  
\usepackage{helvet} 
\usepackage{courier}  
\usepackage[hyphens]{url}  
\usepackage{graphicx} 
\usepackage{graphicx}  
\usepackage{mathrsfs}
\usepackage{amsthm}
\usepackage{amsmath}
\newtheorem{definition}{Definition}[section]

\title{Tensor Graph Convolutional Networks for Text Classification}
\author{Xien Liu$^{*1}$, Xinxin You$\thanks{The first two authors contribute equally to this work.} ^{3,4} $, Xiao Zhang$^{1}$, Ji Wu$^{1,2}$, Ping Lv$^{3,4}$ \\
$^1$Department of Electronic Engineering, Tsinghua University, Beijing 100084, China  \\
$^2$  Institute for Precision Medicine, Tsinghua University, Beijing 100084, China \\
$^3$ Tsinghua-iFLYTEK Joint Lab,  iFlytek Research, Beijing 100084, China  \\
$^4$State Key Laboratory of Cognitive Intelligence, Hefei, Anhui 230088, China  \\
}

\begin{document}
\maketitle
%
%
%
\begin{abstract}
Compared to sequential learning models, graph-based neural networks exhibit some excellent properties, such as ability capturing global information. In this paper, we investigate graph-based neural networks for text classification problem.  A new framework TensorGCN (tensor graph convolutional networks), is presented for this task\footnote{The code will be released at \url{https://github.com/xienliu/tensor-gcn-text-classification-tensorflow}}. A text graph tensor is firstly constructed to describe semantic, syntactic, and sequential contextual information.  Then, two kinds of propagation learning perform on the text graph tensor. The first is intra-graph propagation used for aggregating information from neighborhood nodes in a single graph. The second is inter-graph propagation used for harmonizing heterogeneous information between graphs. Extensive experiments are conducted on benchmark datasets, and the results illustrate the effectiveness of our proposed framework.  Our proposed TensorGCN presents an effective way to harmonize and integrate heterogeneous information from different kinds of graphs. 
\end{abstract}
%
%
%
%
 \begin{figure*}[hbt]
 \centering
 \includegraphics[height=5.0cm, width=18cm]{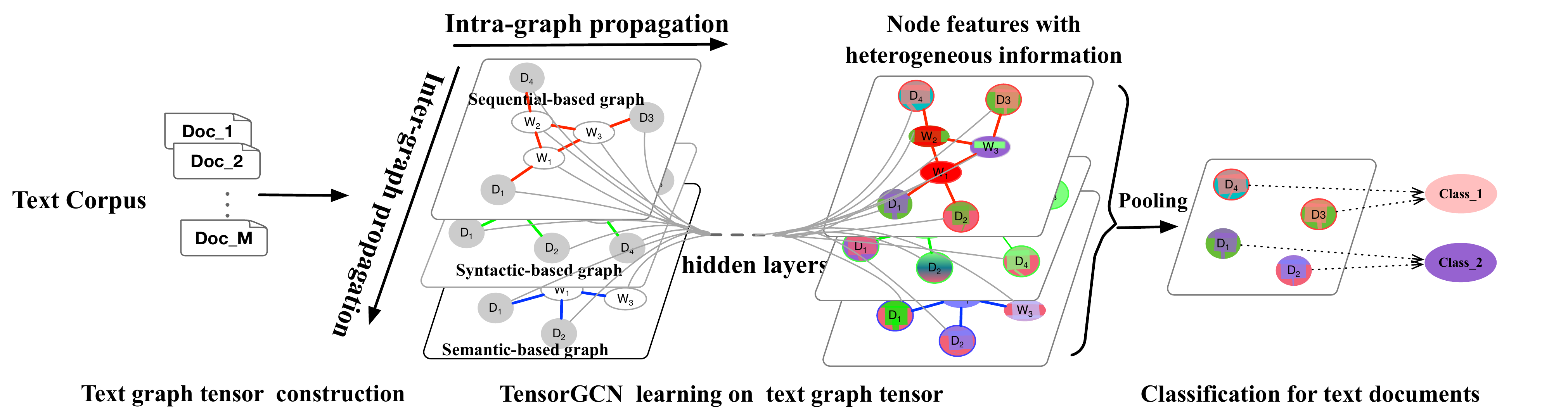}
 \caption{The whole framework of our propsoed TensorGCN for text classification.}
 \label{fig:text-tgcn}
 \end{figure*}
\section{Introduction} 

Text classification is one of the most fundamental tasks in the natural language
processing community. It can be simply formulated as { $\mathcal X \to y$}, where { $\mathcal X$} is a piece of text (such as a sentence/document)
, and { $y \in [0,1]^n$} is the corresponding label vector. In this study, we only consider one label classification problem. In practice,  numerous real applications can be cast into a text classification problem, such as document organization, news filtering, spam detection, EHR-based disease diagnoses \cite{lipton2015learning,che2015deep,miotto2016deep}. 

 Text representation learning is  the first and essential step for the text classification problem. The study of text representation can be summarized into two lines: 1) engineering features and 2)  learning features.  In the study of engineering features,  a piece of text is represented with hand-crafted features, such as bag-of-word features, sparse lexical features \cite{wang2012baselines}, and entity-based features \cite{chenthamarakshan2011concept}.  Recently, text features are automatically extracted from the raw text with machine learning models. According to the learning models, the manners of feature learning  can be further cast into two categories: 1) sequential-based learning models and 2) graph-based learning models.  The most used sequential-based learning models include  convolutional neural networks \cite{kim2014convolutional,zhang2015character,conneau2016very} and recurrent neural networks \cite{liu2016recurrent,tai2015improved}, which promote to capture text features from local consecutive word sequences. Recently, graph-based learning models, such as graph neural networks \cite{cai2018comprehensive,battaglia2018relational}, have attracted widespread attention and been successfully applied for solving the text classification problem \cite{kipf2016semi,yao2019graph}. 

Different from sequence learning models, graph neural networks (GNN) can directly deal with complex structured data, and prioritize exploiting global features. Data from many real applications can be naturally cast into a graph \cite{deac2019drug,xu2019mr}, but it does not hold for sequential free text.  Therefore, GNN-based text learning must include two stages: 1) construct graphs from free text, and 2) build a graph-based learning framework. A straightforward manner of graph construction is to build and connect relationships between words/entities in the free text. Recently, Yao et al. \shortcite{yao2019graph} proposed a text graph-based neural network TextGCN, which achieved state-of-the-art performance in some benchmark datasets of text classification. 
In their framework, a text graph is firstly constructed based on the sequential contextual relationship between words.  Then a graph convolutional network (GCN) \cite{kipf2016semi} is employed to learn on the text graph. According to the more recently reported in \cite{luo2016bridging,bastings2017graph,vashishth2019incorporating}, more contextual information should be considered, such as semantic and syntactic contextual information.

Inspired by the recent progress, we propose a new graph-based text classification framework TensorGCN (see figure \ref{fig:text-tgcn}). 
Semantic-based, syntactic-based, and sequential-based text graphs are firstly constructed to form a text graph tensor. The graph tensor is used to capture text information of semantic context, syntactic context, and sequential context, respectively. As pointed out in \cite{vashishth2019incorporating},  the jointly learning of multi-graphs is a very challenging task, especially when the graphs hold very heterogeneous information.  To encode the heterogeneous information from multi-graphs,  TensorGCN simultaneously performs two kinds of propagation learning.  For each layer, an intra-graph propagation is firstly performed  for  aggregating information from neighbors of each node.  Then an inter-graph propagation is used to harmonize heterogeneous information between graphs.  Experimental results on some benchmark datasets illustrate the effectiveness of TensorGCN for the text classification problem.  The main contributions are summarized as follows:
\begin{itemize}
	\item A text graph tensor is constructed to describe contextual information with semantic, syntactic, and sequential constraints, respectively. 
	\item A learning method TensorGCN is proposed to harmonize and integrate heterogeneous information from multi-graphs.	
	\item Extensive experiments are conducted on several benchmark datasets to illustrate the effectiveness of TensorGCN for text classification. 
\end{itemize} 
%
%
%
\section{Related work}
Recently, graph neural networks have received growing
attentions and successfully used in many applications \cite{tu2019multi,yao2019graph,cao2019multi,vashishth2019incorporating,bastings2017graph,kipf2016semi,li2018deeper,xu2019mr}. Yao et al. \shortcite{yao2019graph} employed the standard graph convolutional networks \cite{kipf2016semi} for text classification. 
 In their work, only one text graph is used to describe the local co-occurring constraint; non-sequential text information has not been fully considered. More recently,  Vashishth et al. \shortcite{vashishth2019incorporating} utilized graph convolutional networks to incorporate syntactic/semantic information for word embedding learning. In their work, the semantic relationship between words is established based on additional semantic sources, which will restrict its application. They used  syntactic and semantic information independently, without jointly learning on the semantic and syntactic graphs. Cao et al. \shortcite{cao2019multi} presented a multi-channel graph learning framework to align entities. In their framework, two graphs are "crudely"  forced to share the trainable parameters in the learning process.  When graphs are very different and  heterogeneous, the parameter-sharing strategy does not work. 
%
%
%
\section{Methods}
In this study, we utilize graph convolutional networks (GCN) as a base component for text graph tensor learning, due to its simplicity and effectiveness in practice.  In this section, we firstly give a brief overview of GCN and present a simple definition of graph tensor. Then, we introduce details of how to construct a graph tensor from a text corpus. Ultimately, we present the TensorGCN learning model. 
%
\subsection{Graph convolutional networks (GCN)} 
A GCN is a generalization version of the traditional convolutional neural networks (CNN)， which can operate directly on a graph.  Formally, consider a  graph { $G=(V, E, A)$}， where  { $V$($|V| = n$)} is the set of graph nodes, $E$ is the set of graph edges,  and $A\in R^{n\times n}$ is the graph adjacency matrix. In GCN learning,  hidden layer representations are obtained by encoding both graph structure and features of nodes with a kind of propagation rule
%
\begin{equation} 
	H^{(l+1)} = f(H^{(l)}, A), \  \   \ (l=0,1,\cdots, L),
\end{equation}
where { $H^{(l)} \in R^{n \times d_l}$} is the feature matrix of  the $l$th layer ($d_l$: number of features for each node in the $l$th layer) and $L$ is the number of layers of GCN. A commonly used layer-wise propagation rule is 
%
\begin{equation} 
\label{equ:gcn}
	H^{(l+1)} = f(H^{(l)}, A) = \sigma(\hat{A} H^{(l)} W^{(l)}),
\end{equation}
where { $\hat{A} = \tilde{D}^{-\frac{1}{2}} \tilde{A} \tilde{D}^{-\frac{1}{2}} $} is a symmetric normalization of the self-connections added adjacency matrix { $\tilde{A} = A + I $ }({ $I$} : an identity matrix), {  $\tilde{D}$ } is the diagonal node degree matrix with { $\tilde{D}(i, i) = \sum_{j} \tilde{A}(i, j)$ , $W^{(l)} \in R^{d_l \times d_{l+1}}$} is a layer-specific trainable weight matrix, where { $d_l$ } and { $d_{l+1}$} are the number of features for each node in the { $l$}th layer and the { $(l+1)$} th layer, respectively.    
{ $\sigma$ } is a  non-linear activation function, such as  { ReLU} or { Leaky ReLU}. \emph{ When in the last layer { $\sigma$ } is always  set to be the  softmax function, and the number of features for each node is equal to the number of labels.} Specifically,  {\bf { $H^{(0)}$}  is the initial feature matrix}, where each row represents a node's initial input feature. 
%
\subsection{Graph tensor definition}
\label{sec:graph-tensor}
Since we want to utilize a series of graphs to fully investigate our interested data (e.g. text documents) and different graphs represent different properties of the data. All of these graphs are packed into a graph tensor. For convenience of study, here we make a formal definition of this kind of graph tensor, which  consists of multiple graphs sharing the same nodes.
%
\begin{definition}
\label{def:graph-tensor}
	  { $\mathcal G$} is a {\bf graph tensor}, where { $\mathcal G = ( G_1, G_2, \cdots, G_r) $ } and  { $ G_i = (V_i, E_i, A_i)$ } , if { $V_i = V_j$} ({ $i,j = 1,2,\cdots, r$} ) and { $A_i \neq A_j$ } (when { $i \neq j$} ). 
\end{definition}
Where { $G_i$ } is the $i$th graph in the graph tensor { $\mathcal G$} ,  { $V_i$ ($|V_i| = n$) }  is the set of the $i$th graph nodes, and { $E_i$} is the set of the $i$th graph edges.  { $A_i$} is the $i$th graph adjacency matrix. Since a graph structure is mainly described with its adjacency matrix, for convenience we also pack the adjacency matrices into a tensor. { $\mathcal A = ({A}_1, {A}_2, \cdots, {A}_r ) \in R ^{r\times n \times n}$ } is a {\bf graph adjacency tensor} , where { ${A}_i$ $ (i=1,2, \cdots, r)$ } is the adjacency matrix of the $i$th graph in the graph tensor { $\mathcal G$ }.  The graph feature matrix in the formula (\ref{equ:gcn}) is generalized into a {\bf graph feature tensor} { $\mathcal H^{(l)} = ({H}_1^{(l)}, {H}_2^{(l)}, \cdots, {H}_r^{(l)}) \in R ^{r\times n \times d_l}$ }, where { ${H}_i^{(l)}\in R^{n \times d_l}$ $ (i=1,2, \cdots, r)$} is the  feature matrix of the $i$th graph in { $\mathcal G$}. When $l=0$, the graph feature tensor { $\mathcal H^{(0)}$} denotes the initialized input features. 
%
%
\begin{figure*}[tbh]
	\centering
	\includegraphics[height=3cm, width=16cm]{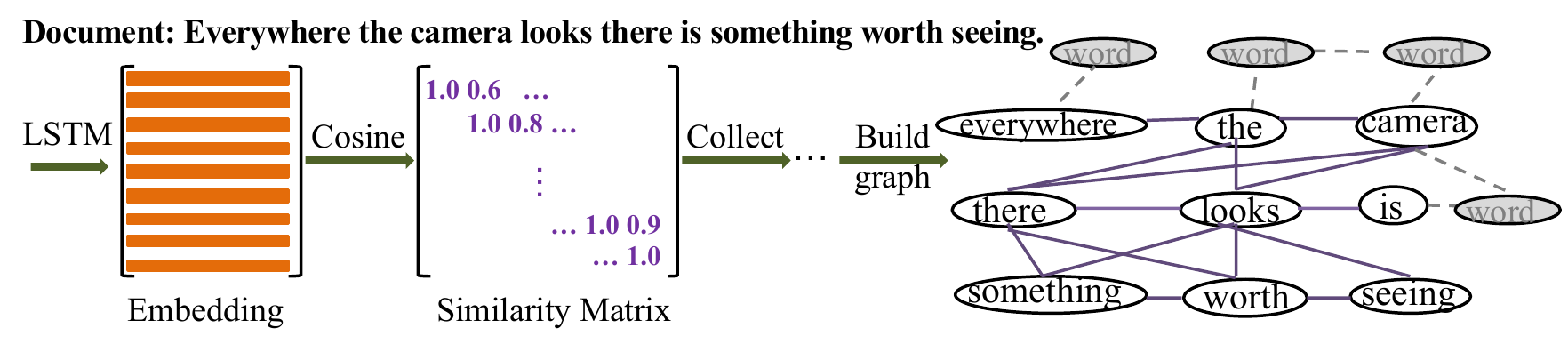}
    \caption{An example to show how to build relationship between words with LSTM encoded semantic information. Take one document for example, the semantic-based graph is constructed by collecting all semantic relationship word pairs over all the text corpus. For details see section \emph{Text Graph Tensor Construction: Syntactic-based graph.} }
  \label{fig:semantic-graph}
\end{figure*}
%
%
\subsection{Text graph tensor construction}
In this section, we describe how to construct a meaningful graph tensor to describe text documents at different knowledge/language properties. A straightforward manner of constructing a graph from text is to treat words and documents as nodes in the graph. Therefore, we need to build two kinds of edges between nodes: word-document edges, and word-word edges.  \emph{The word-document edges are built based on word occurrence in documents and the edge weights are measured with  the term frequency-inverse document frequency (TF-IDF) method.  If no otherwise specified, in all the following graphs  the word-document edges are built and measured with TF-IDF.}  In this study, we build word-word edges based on three different language properties: semantic information, syntactic dependency, and local sequential context. Based on these different kinds of word-word edges, we construct a series of text graphs to depict text documents. Details are presented in the following sections. 
%
\subsubsection{Semantic-based graph}
Motivated  by LSTM shows potential ability to capture semantic information for word representation \cite{iacobacci2019lstmembed}, we propose a LSTM-based method to construct a semantic-based graph from text documents (see figure \ref{fig:semantic-graph}). There are three main steps: 
{ 
\begin{itemize}
	\item [-] Step 1:  Train a LSTM on the {\bf training data} of the given task (e.g. text classification here). 
	\item [-] Step 2:  Get semantic features/embeddings with LSTM for all words in each document/sentence of the corpus. 
	\item [-] Step 3:  Calculate word-word edge weights based on word semantic embeddings over the corpus. 
\end{itemize}
}
For every sentence/document, we obtain word semantic features/embeddings from the outputs of the trained LSTM and calculate cosine similarity between words. If the similarity value exceeds a predefined threshold of $\rho_{sem}$, it means that the two words have a semantic relationship in the current sentence/document.  We count the number of times for each pair of words having a semantic relationship over the whole corpus. The edge weight of each pair of words (nodes in the semantic-based graph) can be obtained by
\begin{equation} 
\label{equ:semantic}
	d_{\text{semantic}}(w_i, w_j) = \frac{\# N_{\text{semantic}}(w_i, w_j) }{\# N_{\text{total}}(w_i, w_j)},
\end{equation}
where { $d_{\text{semantic}}(w_i, w_j)$ } denotes the edge weight between words $w_i$ and $w_j$, { $\# N_{\text{semantic}}(w_i, w_j)$ } is the number of times that the two words have semantic relationship over all sentences/documents in the corpus,  and { $\# N_{\text{total}}$} is the number of times that the two words exist in the same sentence/document over the whole corpus. 
%
\subsubsection{Syntactic-based graph} For each sentence/document in the corpus, we first utilize Stanford CoreNLP parser to extract the dependency between words. Though the extracted dependency is directed, for simplicity, we treat it as an undirected relationship. Similar to the strategies used in the above  semantic graph, we count the number of times for each pair of words having syntactic dependency over the whole corpus and calculate the edge weight of each pair of words (nodes in the syntactic-based graph) by 

\begin{equation} 
\label{equ:syntactic}
	d_{\text{syntactic}}(w_i, w_j) = \frac{\# N_{\text{syntactic}}(w_i, w_j) }{\# N_{\text{total}}(w_i, w_j)},
\end{equation}
where { $d_{\text{syntactic}}(w_i, w_j)$ } denotes the edge weight between words $w_i$ and $w_j$, { $\#  N_{\text{syntactic}}(w_i, w_j)$ } is the number of times that the two words have syntactic dependency relationship over all sentences/documents in the corpus,  and {  $\# N_{\text{total}}$} , as used in the formula (\ref{equ:semantic}), is the number of times that the two words exist in the same sentence/document over the whole corpus. 

\subsubsection{Sequential-based graph} 
Sequentia context depicts the local co-occurrence (between words) language property, which has been widely used for text representation learning. 
In this study, we utilize point-wise mutual information (PMI) to describe this kind of sequence context information using a sliding window strategy.  The edge weight of each pair of words (nodes in the sequential based graph) is calculated by
\begin{equation} 
	d_{\text{sequential}}(w_i, w_j) = \log \frac{p(w_i, w_j)}{p(w_i)p(w_j)},
\end{equation}
where { $p(w_i, w_j)$}  is the probability of the word pair { ($w_i$, $w_j$)} co-occurring in the same sliding window,  which is always estimated by { $\frac{\#N_{\text{co-occurrence}}(w_i, w_j)}{\# N_{\text{windows}}}$, where  $\# N_{\text{windows}} $}  is the total number of the sliding windows over the whole text corpus and { $\#N_{\text{co-occurrence}}(w_i, w_j)$ } is the number of times that the word pair { ($w_i, w_j$) } co-occurs in the same sliding windows over the whole text corpus. { $p(w_i) = \frac{\# N_{\text{occurrence}(w_i)}}{\# N_{\text{windows}}}$ } is the probability that the word {  $w_i$}  is occurring in a given window over text corpus, where { $\# N_{\text{occurrence}(w_i)}$ } is the number of times that the word { $w_i$ } occurs in the sliding windows over the whole text corpus. 

\subsection{Graph tensor learning}
After obtaining a text graph tensor, we focus on exploiting effective learning frameworks to perform GCN on the graph tensor. In the following, we will first introduce a preliminary model ``Merge edges + GCN" and then propose the TensorGCN model, which can directly learn on a graph tensor in a straightforward but effective way. 

\subsubsection{Preliminary model: merge edges + GCN} A straightforward manner of dealing with this problem is to firstly reduce the graph tensor into a single graph and then utilize the standard GCN learner (\ref{equ:gcn}) to perform the learning process on the single graph.  As we discussed in the above {\emph definition \ref{def:graph-tensor}}, here we mainly focus on the kind of graph tensor in which all the graphs share the same set of nodes, edges are the only difference. Therefore, we only need to merge the edges into one graph by pooling the adjacency tensor   { $pooling(\mathcal A) =pooling({A}_1, {A}_2, \cdots, {A}_r)$} , such as $max \ pooling$, or $mean \ pooling$. A series of preliminary experiments illustrate that it does not work to use  $max  \ pooling$ or $mean \  pooling$ directly, due to that the graphs in the tensor are very heterogeneous and edge weights from different graphs do not match. Therefore, we employ a simple \emph{edge-wise attention} strategy to harmonize edge weights from different graphs.  The adjacency matrix of the merged graph is { $ A_{\text{merge}} = pooling(\mathcal A) = \sum_{i=1}^{r} W_{\text{edge}}^i \odot  A_i$},  where { $ W_{\text{edge}}^i$} is the edge-wise attention matrix with the same size of the adjacency matrix, and $\odot$ is the matrix dot product. 
     \begin{figure*}[hbt]
	\centering
	\includegraphics[height=5.0cm, width=18cm]{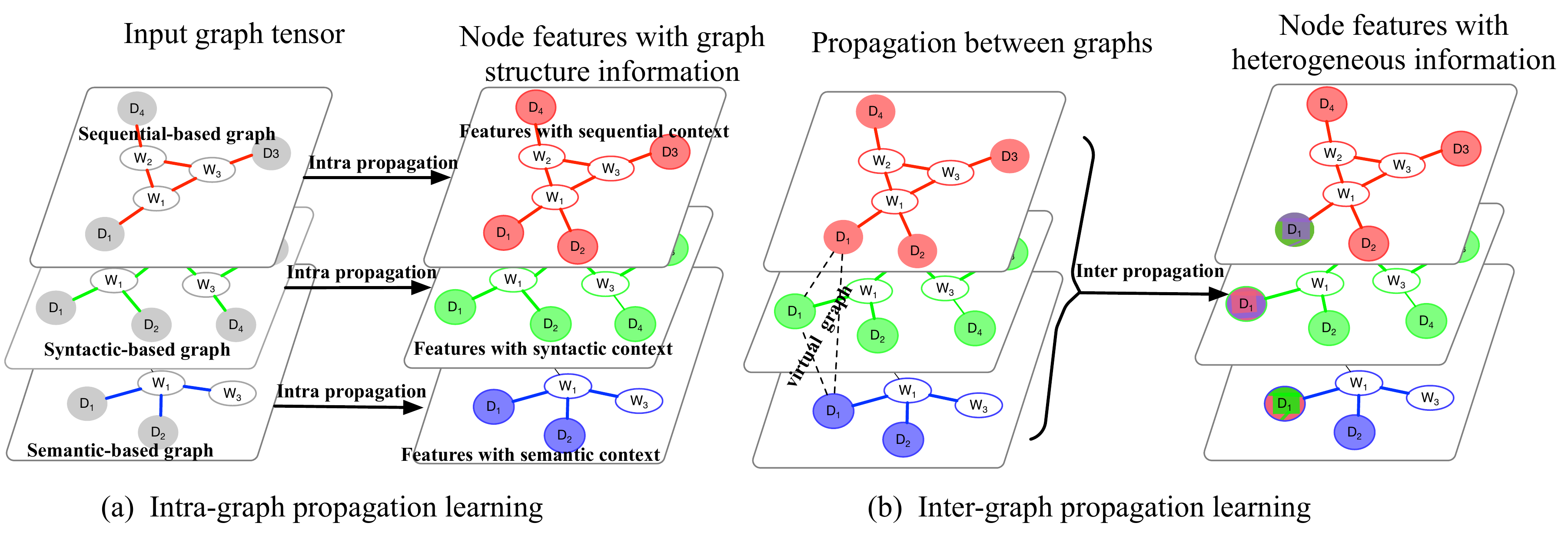}
    \caption{Take the text graph tensor (see  Section  \emph{text Graph Tensor Construction} ) with three word nodes and three document nodes for example to show one of the layer in the TensorGCN learning process. (a): Intra-graph propagation learning from the input graph tensor; (b): Inter-graph propagation learning with using the output of the intra-graph propagation as its input. Here we just take one virtual graph as an example to show how to harmonize heterogeneous information with inter-graph propagation learning. In practice, all virtual graphs have to perform inter-graph propagation learning.}
    \label{fig:tgcn}
    \end{figure*}
\subsubsection{TensorGCN}   The above preliminary model takes a ``rude" manner that treats all graphs in the same representation space and squeezes them into one graph, to some degree destroying the structure of the tensor. Different graphs maintain different properties to depict the given data. Therefore, it's necessary to release some degree of freedom to the learning of different graphs in the tensor. On the other hand, we need to design a mechanism to harmonize heterogeneous information and structures of different graphs during the learning process.  The underlying principle of graph neural network learning is that nodes pass message (coordinate information with each other) and update their feature representation by propagating information among other nodes within the neighborhood.  
    Motivated by this  message passing schema, we utilize a similar manner to propagate  information between different graphs and thus generalize the single graph neural network learning formula (\ref{equ:gcn}) into a graph tensor version TensorGCN, which can directly perform convolutional learning on a tensor graph. 
      For each layer of TensorGCN, we perform two kinds of propagation learning:  first \emph{intra-graph propagation}  and  then \emph{inter-graph propagation} (see figure \ref{fig:tgcn}). We take the $l$th layer of TensorGCN for example.  
  \begin{equation}  
\mathcal H^{(l)} \stackrel{f_{\text{intra}}} \longrightarrow \mathcal H^{(l)}_{\text{intra}}  \stackrel{f_{\text{inter}}} \longrightarrow \mathcal H^{(l+1)},
\end{equation}
where { $\mathcal H^l \in R^{r\times n \times d_l}$}  ($n$ : number of nodes, $d_l$: feature dimension in the $l$th layer), as defined in the section of \emph{graph tensor definition}, is the hidden feature tensor of the $l$th layer in TensorGCN, and { $\mathcal H_{\text{intra}}^{(l)}$} is the output feature tensor after performing intra-graph propagation. { $f_{\text{inra}}$}  and { $f_{\text{inter}}$} denote the intra-graph propagation and inter-graph propagation, respectively.  The learning details of the two kinds of propagation are described next.
 
The \textbf {intra-graph propagation} learning is to aggregate information from neighbors of each node within  a graph (see figure \ref{fig:tgcn}-(a)).  Therefore, the learning schema is almost identical to formula (\ref{equ:gcn}); the only difference is that all graphs have to perform GCN learning, resulting in a tensor version. 
  Given the graph adjacency tensor { $\mathcal A = ({A}_1, {A}_2, \cdots, {A}_r) \in R ^{r\times n \times n}$}, the feature of the $i$th graph in the $l$th layer is updated by the intra-graph propagation { $f_{\text{intra}}$} as follows
\begin{equation} 
\label{equ:tgcn-intra}
	\mathcal H^{(l)}_{\text{intra}}(i,:,:) = \sigma( \hat{\mathcal A}(i,:, :)  \mathcal H^{(l)}(i,:,:)  W^{(l,i)}_{\text{intra}}),
\end{equation}
where { $\hat {\mathcal{A}} \in R ^{r\times n \times n}$ } is the normalized symmetric graph adjacency tensor consisting of a series of normalized symmetric adjacency matrix. For example,  { $\hat{\mathcal A}(i,:, :) = \tilde {\mathcal {D}}(i,:,:)^{-\frac{1}{2}} \tilde{\mathcal A}(i,:,:) \tilde {\mathcal {D}}(i,:,:)^{-\frac{1}{2}}$  } is a symmetric normalization of the $i$th graph adjacency matrix, where { $\tilde{\mathcal{A}} = \mathcal A + \mathcal I$ ( $\mathcal I$ } is an identity tensor consisting of $r$ identity matrix) and { $ \mathcal {D}$}  is the node degree tensor consisting of  $r$ diagonal node degree matrices { $ (D_1, D_2, \cdots, D_r)$}. Unlike formula (\ref{equ:gcn}), here the trainable weight matrix is designed to be layer and graph specific. For example, { $ W^{(l,i)}_{\text{intra}} $} is the weight matrix of the $i$th graph at $l$th layer \footnote{We experimented with the setting that all graphs share the same trainable weights, but our preliminary experiments illustrate that this setting does not work. It should be noticed that the trainable weight is the only variable parameters of the GCN learner. It's not reasonable to roughly compel all graphs to share this only variable parameters, especially when graphs are strong heterogeneous. 
    }.  $\sigma$, as defined in the formula (\ref{equ:gcn}), is a kind of activation function. 

 The \textbf {inter-graph propagation} learning is to propagate/exchange information between different graphs in the tensor (see the figure \ref{fig:tgcn}-(b)), such that the heterogeneous information from the different graphs can be gradually fused into an accordant one.  To achieve this purpose, we construct a series of  special graphs, called  \textbf{virtual graphs}, by connecting with nodes across the graphs in the tensor. As defined in the \emph {definition \ref{def:graph-tensor}} all graphs in the tensor actually  share the same set of nodes, that is { $V_i = V_j$  $(i, j = 1, 2, \cdots, n)$}. We make the ``copy nodes"   { $V_i(1),V_i(2),\cdots, V_i(r) $}  (actually they are the same node) from different graphs connect to each other. Ultimately, we in total obtain $n$  virtual graphs, resulting in a \textbf{ new graph adjacency tensor { $\mathcal A^{+} \in R^{r \times r \times n}$ }} by collecting edge weights (specially here all weights are 1) of the $n$ virtual graphs. The inter-graph information propagation learning { $f_{\text{inter}}$ } on the virtual graphs is  carried out by
 \begin{equation} 
	\mathcal H^{(l+1)}(:,j,:) = \sigma( {\mathcal A}^{+}(:,:, j)  \mathcal H_{\text{intra}}^{(l)}(:,j,:)  W^{(l,j)}_{\text{inter}}),
\end{equation}
 where { $\mathcal H^{l+1} \in R^{r \times n \times d_{l+1}}$} is the output of inter-graph propagation, also the input feature tensor of the $l+1$th layer in TensorGCN.  { $ W^{(l,j)}_{\text{inter}}$ } is the trainable weight in the inter-graph propagation learning. It's important to be noted that here the new adjacency matrix { $\mathcal A^{+}(:,:, j)$}  are used neither doing symmetric normalization nor adding self-connections, which is very different from the formula (\ref{equ:gcn}) and (\ref{equ:tgcn-intra}). It's not necessary to do renormalization to the adjacency matrix of the virtual graphs since all nodes in a virtual graph are connected to each other, and the edge weights are set to 1.  The self-connection strategy (always used in the standard GCN) is intentionally not used here for the purpose that heterogeneous information is more effectively fused together

 In the last layer of TensorGCN, after completed inter-graph propagation, we perform a mean pooling over graphs to obtain the final representation of document nodes for classification. 
\section{Experiments and results analysis}
In this section, we  evaluate the performance of our proposed TensorGCN based text classification framework, then carefully examine the effectiveness of our constructed text graph tensor and the ability of our developed TensorGCN algorithm for joint learning on multi-graphs.

\subsection{Symbol \& abbreviation}
For convenience, some symbols and abbreviations are used in the following experiments, which are listed in the table \ref{tab:symbol}.

\begin{table}[tbh]
\centering
\caption{Symbol \& abbreviation}
	\begin{tabular}{c|c}
	\hline
	Symbol \& Abbreviation  & Explanation  \\
     \hline
     20NG   &  The 20-Newsgroups dataset \\
      R8   & R8 Reuters dataset \\
      R52 &  R52 Reuters dataset\\
      Ohsumed &  Ohsumed dataset \\
      MR  & Movie Review dataset \\
      \hline
      SemGraph  & Semantic based graph  \\ 
      SynGraph  & Syntactic based graph  \\ 
      SeqGraph   & Sequential based graph \\ 
      \hline
      SemEdges & Edges of SemGraph  \\
      SynEdges  & Edges of SynGraph \\
      SeqEdges   & Edges of SeqGraph \\
      \hline
	\end{tabular}
\label{tab:symbol}
\end{table}

\begin{table*}[htb]
\centering
\caption{Summary statistics of datasets and the number of edges in the constructed text graph tensor.}
	\begin{tabular}{@{}c|cccccc|ccc@{}}
    \hline
	\bf{Dataset} & \bf{\# Doc} &  \bf{\# Train}	&  \bf{\# Test}	 &  \bf{\# Words} &  \bf{\# Classes}   &  \bf{Avg Length}  & \bf {\#SemEdges} & \bf{\# SynEdges} & \bf{\# SeqEdges} \\
	\hline
	20NG&18,846&11,314&7,532&42,757&20&221.26 & 87,431,640 & 13,816,858 & 	22,413,246 \\
R8&7,674&5,485&2,189&7,688&8&65.72 & 6,472,304  & 408,772 & 2,841,760  \\
R52&9,100&6,532&2,568&8,892&52&69.82  & 8,204,096  & 553,218 &  3,574,162 \\
Ohsumed&7,400&3,357&4,043&14,157&23&135.82 & 7,442,388  & 989,102 & 6,867,490  \\
MR&10,662&7,108&3,554&18,764&2&20.39 & 1,660,038 & 177,294 & 1,504,598  \\
	\hline
	\end{tabular}
\label{tab:dataset}
\end{table*}
\begin{table*}[htb] 
\centering
\caption{Test accuracy comparison with baselines on benchmark datasets.}
	\begin{tabular}{@{}c|ccccc@{}}
    \hline
	\bf{ Model } & \bf{20NG} &  \bf{R8}	&  \bf{R52}	 & \bf{Ohsumed}  &	\bf{MR}	 \\
\hline
TF-IDF + LR &	0.8319 $\pm$ 0.0000&0.9374 $\pm$ 0.0000&0.8695 $\pm$ 0.0000&0.5466 $\pm$ 0.0000&0.7459 $\pm$ 0.0000  \\
\hline
CNN-rand& 0.7693 $\pm$ 0.0061&0.9402 $\pm$ 0.0057&0.8537 $\pm$ 0.0047&0.4387 $\pm$ 0.0100&0.7498 $\pm$ 0.0070 \\
CNN-non-static&	0.8215 $\pm$ 0.0052&0.9571 $\pm$ 0.0052&0.8759 $\pm$ 0.0048&0.5844 $\pm$ 0.0106&0.7775 $\pm$ 0.0072 \\
LSTM&	0.6571 $\pm$ 0.0152&0.9368 $\pm$ 0.0082&0.8554 $\pm$ 0.0113&0.4113 $\pm$ 0.0117&0.7506 $\pm$ 0.0044 \\
LSTM (pretrain)	& 0.7543 $\pm$ 0.0172&0.9609 $\pm$ 0.0019&0.9048 $\pm$ 0.0086&0.5110 $\pm$ 0.0150&0.7733 $\pm$ 0.0089  \\
Bi-LSTM &	0.7318 $\pm$ 0.0185&0.9631 $\pm$ 0.0033&0.9054 $\pm$ 0.0091&0.4927 $\pm$ 0.0107&0.7768 $\pm$ 0.0086 \\
\hline
PV-DBOW	& 0.7436 $\pm$ 0.0018&0.8587 $\pm$ 0.0010&0.7829 $\pm$ 0.0011&0.4665 $\pm$ 0.0019&0.6109 $\pm$ 0.0010 \\
PV-DM&	0.5114 $\pm$ 0.0022&0.5207 $\pm$ 0.0004&0.4492 $\pm$ 0.0005&0.2950 $\pm$ 0.0007&0.5947 $\pm$ 0.0038 \\
fastText &	0.1138 $\pm$ 0.0118&0.8604 $\pm$ 0.0024&0.7155 $\pm$ 0.0042&0.1459 $\pm$ 0.0000&0.7217 $\pm$ 0.0130  \\
fastText (bigrams) &	0.0734 $\pm$ 0.0068&0.8295 $\pm$ 0.0003&0.6819 $\pm$ 0.0004&0.1459 $\pm$ 0.0000&0.6761 $\pm$ 0.0279  \\
SWEM &	0.8516 $\pm$ 0.0029&0.9532 $\pm$ 0.0026&0.9294 $\pm$ 0.0024&0.6312 $\pm$ 0.0055&0.7665 $\pm$ 0.0063 \\
LEAM &	0.8191 $\pm$ 0.0024&0.9331 $\pm$ 0.0024&0.9184 $\pm$ 0.0023&0.5858 $\pm$ 0.0079&0.7695 $\pm$ 0.0045  \\
\hline
Graph-CNN-C &	0.8142 $\pm$ 0.0032&0.9699 $\pm$ 0.0012&0.9275 $\pm$ 0.0022&0.6386 $\pm$ 0.0053&0.7722 $\pm$ 0.0027  \\
Graph-CNN-S &	$-$  & 0.9680 $\pm$ 0.0020&0.9274 $\pm$ 0.0024&0.6282 $\pm$ 0.0037&0.7699 $\pm$ 0.0014  \\
Graph-CNN-F&	$-$  & 0.9689 $\pm$ 0.0006&0.9320 $\pm$ 0.0004&0.6304 $\pm$ 0.0077&0.7674 $\pm$ 0.0021  \\
Text GCN &	0.8634 $\pm$ 0.0009&   0.9707 $\pm$ 0.0010&0.9356 $\pm$ 0.0018&0.6836 $\pm$ 0.0056&0.7674 $\pm$ 0.0020 \\
 TensorGCN & {\bf 0.8774 $\pm$ 0.0005} & {\bf 0.9804 $\pm$  0.0008}  &  \bf{0.9505 $\pm$ 0.0011}  &  \bf{0.7011 $\pm$ 0.0024} &  \bf{0.7791 $\pm$ 0.0007} \\
	\hline
	\end{tabular}
\label{tab:compare-total}
\end{table*}

\subsection{Datasets and baselines}
A suite of recently widely used benchmark datasets were used to perform experiments and analysis. The benchmark corpora consist of five text classification datasets: 20-Newsgroups dataset, Ohsumed dataset, R52 Reuters dataset, R8 Reuters dataset, and Movie Review dataset. These datasets involve many life genres, such as movie reviews, medical literature, and news document, etc. The  Movie Review dataset is designed for binary sentiment classification. The  20-Newsgroups dataset, R52 Reuters dataset, R8 Reuters dataset are news classification. The Ohsumed dataset is medical literature. A summary statistics of the benchmark datasets is presented in table \ref{tab:dataset}. 

The baselines  can be categorized into four categories: 1) powerful traditional models, such as  TF-IDF + LR ; 2) word embedding based models, mainly including  PV-DBOW \cite{le2014distributed}, PV-DM \cite{le2014distributed}, fastText \cite{joulin2016bag}, and some recent  state-of-the-art  methods such as SWEM \cite{shen2018baseline} and LEAM \cite{wang2018joint}; 3) sequence deep learning models which use CNN \cite{kim2014convolutional}, LSTM \cite{liu2016recurrent} or Bi-LSTM as feature extractor; 3) graph based representation learning models, such as Graph-CNN-C \cite{defferrard2016convolutional}, Graph-CNN-S \cite{bruna2013spectral}, Graph-CNN-F \cite{henaff2015deep}, and text GCN \cite{yao2019graph}. More detail descriptions about baselines and datasets can be found in \cite{yao2019graph}. To conduct a fair comparison study, we  utilize the results of baselines reported in \cite{yao2019graph}, and use the same datasets with the same settings to test our model.

\subsection{Experiment settings}
 In the construction of the sequential-based graph, the window size is 20 as used in \cite{yao2019graph}, and in the semantic-based graph construction, the initial word embeddings are pre-trained with Glove, and the dimension is 300, the dimension of LSTM is also set to 300.
As suggested and used in \cite{yao2019graph}, in this study, we also take a two-layers TensorGCN and dimension of the node embedding in the first layer is 200, and the dimension of the node embedding in the second layer is equal to the number of the labels. In the training process, the dropout rate is 0.5, and $L_2$ loss weight is 5e-6. As used in \cite{yao2019graph}, we randomly selected 10\% of the training set as the validation set, which labels are not be used for training. A maximum of 1000 epochs and  Adam optimizer with a learning rate of 0.002 are used, and early stopping is performed when validation loss does not decrease for ten consecutive epochs. 

 All results reported in this study are the mean values of ten  independent runs.  Due to space limitations, the standard deviation values are given only in the table \ref{tab:compare-total}.

\begin{figure*}[b!th]
	\centering
	\includegraphics[height=4.0cm, width=16cm]{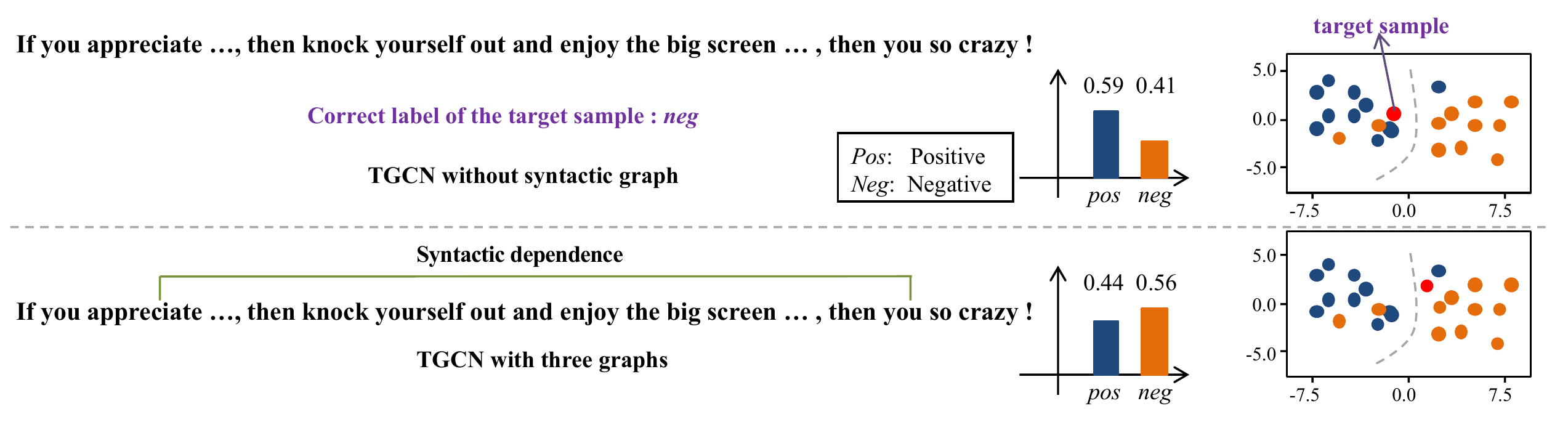}
	\setlength{\abovecaptionskip}{0pt}
    \caption{An example from the MR dataset to show how the important syntactic dependency information has been  used by TensorGCN to enhance sentiment classification performance.}
  \label{fig:case-syntactic}
\end{figure*}

\subsection{Results analysis}
\subsubsection{Performance}  A comprehensive experiment is conducted on the benchmark datasets. The results presented in the table \ref{tab:compare-total} show that our proposed TensorGCN significantly outperforms all baselines (including some state-of-the-art embedding learning and graph-based models). It should be noted that the strong baseline Text GCN is equivalent to TensorGCN with only using the sequential-based graph.  
\begin{table}[h!tb]
\centering
\caption{Analysis of the effectiveness of SeqGraph, SynGraph, and SemGraph. In the upper half table, the first three rows are the results of only using a single graph, while the lower half table is the results of ablation experiments. For example, SemGraph(w/o) means TensorGCN without using SemGraph. Graph tensor means three graphs are used. }
	\begin{tabular}{@{}c@{}c@{}c@{}c@{}c@{}c@{}}
    \hline
	\; \bf{ Model } \; & \;\, \bf{20NG} \;\, & \;\;\,\bf{R8}\;\; & \;\;\;\;\bf{R52}\;\;\; &\bf{Ohsumed} &\bf{MR} \\
\hline
SeqGraph  &  0.8634 & 0.9707  & 0.9356 & 0.6836     & 0.7674 \\
SynGraph  &  0.8478 & 0.9534  &0.9299  &0.6658 & 	0.7496 \\  
SemGraph  & 0.8517  & 0.9520 &	0.9245 &  	0.6626	   & 0.7620   \\
\hline
SemGraph(w/o)  & 0.8688 &  0.9776 &	0.9334	& 0.6917	 & 0.7763 \\
SynGraph(w/o)  & 0.8707  & 0.9753	& 0.9529 & 	0.6854 & 	0.7710 \\
SeqGraph(w/o) & 0.8699 & 0.9753 & 	0.9486  &	0.6921  &	0.7746 \\
{Graph tensor} & {\bf 0.8774} & {\bf 0.9804}  &  \bf{0.9505}  &  \bf{0.7011} &  \bf{0.7791} \\
	\hline
	\end{tabular}
\label{tab:graph-analysis}
\end{table}
\subsubsection{Analysis of text graph tensor}
We also examine and perform an analysis of the effectiveness of our constructed text graph tensor. The results of a single text graph, pairs of every two graphs, and all three graphs (graph tensor) are  presented in the table \ref{tab:graph-analysis}. We notice that graph tensor has the best performance, and each pair of graphs always show better performance than any single graph. The phenomenons illustrate that the proposed three kinds of text graphs are complementary to each other.  We also observed that different text graphs play different roles in different datasets.  For example,  in the ablation experiments, there are worse accuracy results on the Ohsumed dataset and MR dataset without using SynGraph.  It means that syntactic dependency plays an important role in those two datasets. An example from the MR dataset has been presented in figure \ref{fig:case-syntactic} to show how syntactic dependency information enhances the classification performance. 
\begin{table}[h!tb]
\small
\centering
\caption{Analysis of the effectiveness TensorGCN learning. ``Merge edges" is our proposed preliminary model which reduce multi-graphs into a single one with an edge-wise  attention strategy (details see the section of \emph{ Graph Tensor Learning}). TensorGCN(intra) means TensorGCN without using inter-graph propagation, only intra-graph propagation used.}
	\begin{tabular}{@{}c@{}c@{}c@{}c@{}c@{}c@{}}
    \hline
	\; \bf{ Model } \,\; & \;\; \bf{20NG} \;\; & \;\;\;\bf{R8}\;\;\; & \;\;\;\;\;\bf{R52}\;\;\;\; &\bf{Ohsumed}\; &\bf{MR} \\
\hline
Merge edges  &0.8721 &  0.9749 &  0.9463 &  0.6951 &  0.7532\\
TensorGCN(intra) &  0.8701 & {\bf 0.9808} & 0.9498 & 0.6723 & 0.7656 \\
TensorGCN & {\bf 0.8774} &  0.9804  &\bf{ 0.9505}  & \bf{ 0.7011 } &  \bf{0.7791} \\
	\hline
	\end{tabular}
\label{tab:tensorgcn-analysis}
\end{table}

\subsubsection{Analysis to TensorGCN learning}
Finally, we examine the effectiveness of TensorGCN learning. We developed two baselines that can deal with the multi-graphs learning problem.  The first is the ``Merge edges" method, which is our proposed preliminary model in the section of \emph{ Graph Tensor Learning}. The second baseline is TensorGCN(intra), which is TensorGCN but without using an inter-graph propagation strategy.  Cao et al. \shortcite{cao2019multi}  also presented a method to deal with multiple graphs. Still, as we discussed above, it does not work (Our preliminary experiments show the performance is even worse than using a single graph) when graphs are very different and heterogeneous.  

 The comparison results presented in table \ref{tab:tensorgcn-analysis} illustrates that  TensorGCN has the best performance for jointly learning on multi-graphs. Compared to TensorGCN(intra), TensorGCN has better performance in almost all test datasets. Therefore the inter-graph propagation strategy (used for harmonizing information from different graphs) is very important for graph tensor learning. 

\section{Conclusions}
In this study, we propose a text graph tensor to capture features from semantic, syntactic, and sequential contextual information. Experimental results illustrate that these different context constraints are complementary and very important for text representation learning. Furthermore, we generalize the graph convolutional networks into a tensor version TensorGCN which can effectively harmonize and integrate heterogeneous information from multi-graphs by the intra-graph and inter-graph propagation simultaneously learning strategy. 

\section{ Acknowledgments}
We thank the anonymous reviewers for their valuable
comments. This work is supported by the National Key Research and Development Program of China (No.2018YFC0116800).

\bibliographystyle{aaai}
\bibliography{tensor-gcn-text-classification.bib} 
\end{document}